\documentclass{article}
\usepackage{ijcai18}

\usepackage{times}
\usepackage{xcolor}
\usepackage{soul}
\usepackage[utf8]{inputenc}
\usepackage[small]{caption}
\usepackage{url}
\usepackage{booktabs}
\usepackage{todonotes}
\usepackage{enumitem}
\usepackage{algorithm}
\usepackage[noend]{algpseudocode}
\usepackage{caption}
\usepackage{amsmath}
\usepackage{amssymb}
\usepackage{multicol}
\usepackage{subcaption}
\usepackage{multirow}
\usepackage[draft]{hyperref}

\newcommand{\idest}{{\it i.e.}}
\newcommand{\exemp}{{\it e.g.}}

\title{LSTM-Based Goal Recognition in Latent Space}

\author{
Leonardo Amado$^{*}$, João Paulo Aires$^{*}$, Ramon Fraga Pereira$^{*}$, \\ \textbf{Maurício C. Magnaguagno$^{*}$, Roger Granada$^{*}$ \and Felipe Meneguzzi\dag} \\
Pontifical Catholic University of Rio Grande do Sul (PUCRS), Brazil \\
Postgraduate Programme in Computer Science, School of Technology \\
$^{*}$\texttt{$\lbrace$leonardo.amado, joao.aires.001, ramon.pereira,}\\ 
\texttt{mauricio.magnaguagno, roger.granada$\rbrace$@acad.pucrs.br} \\
\dag\texttt{felipe.meneguzzi@pucrs.br} \\
}

\begin{document}

\maketitle

\begin{abstract}

Approaches to goal recognition have progressively relaxed the requirements about the amount of domain knowledge and available observations, yielding accurate and efficient algorithms capable of recognizing goals. 
However, to recognize goals in raw data, recent approaches require either human engineered domain knowledge, or samples of behavior that account for almost all actions being observed to infer possible goals.
This is clearly too strong a requirement for real-world applications of goal recognition, and we develop an approach that leverages advances in recurrent neural networks to perform goal recognition as a classification task, using encoded plan traces for training. 
We empirically evaluate our approach against the state-of-the-art in goal recognition with image-based domains, and discuss under which conditions our approach is superior to previous ones. 

\end{abstract}

\section{Introduction}
\label{sec:introduction}

Goal recognition is the task of identifying the intended goal of an agent under observation by analyzing the agent behavior in an environment. 
Initial approaches on goal recognition were based on planning theories, which require a substantial amount of domain knowledge~\cite{Kautz:1986uv}. 
Subsequent approaches have gradually relaxed such requirements using expressive planning and plan-library-based formalisms~\cite{Avrahami-Zilberbrand2005,Geib2007,Meneguzzi2010a,Fagundes2014} as well as achieving different levels of accuracy
and amount of information available in observations required to recognize goals~\cite{Martin:2015ta,Sohrabi:2016uw,Pereira2016,PereiraOrenMeneguzzi2017}.

Recent work on goal recognition in latent space~\cite{amado2018} overcomes this limitation by building planning domain knowledge from raw data and using such domain knowledge on traditional goal recognition techniques~\cite{PereiraOrenMeneguzzi2017} to infer goals from image data.
However, to build this domain knowledge, their approach requires a substantial amount of training data to create a complete PDDL domain. 
In this paper, we try to mitigate this problem by applying a recurrent neural network to solve the task of goal recognition directly rather than to use the training data to generate domain knowledge. 
Our main goal is to reduce the amount of training data necessary to correctly infer the intended goal of an agent by leveraging a Long short-term memory (LSTM) network. 
Long short-term networks \cite{Hochreiter:1997:LSM:1246443.1246450} are capable of solving classification problems by receiving streams of data and returning a class based on the entirety of the data received. 
These streams of data can be used to model the actions of an agent under observation in goal recognition problems, where the class to be recognized by the LSTM network is the agent's goal. 

Our main contributions are twofold. 
First, we develop an end-to-end machine learning technique for goal recognition~\cite[Chapter 1]{ActivityIntentPlanRecogition_Book2014} based on training an LSTM network in Section~\ref{sec:goalRecognitionInLatentSpace}.
Second, we empirically compare the resulting approach with traditional goal recognition approaches~\cite{Ramirez:2009wo,PereiraOrenMeneguzzi2017} in Section~\ref{sec:experiments}, discuss how our approach relates to the current state of the art in Section~\ref{sec:relatedWork}, and in Section~\ref{sec:conclusions}, discuss the trade-offs between using machine learning exclusively or combining traditional techniques with machine learning. 

\section{Background}
\label{sec:background}

%
%


\subsection{Goal Recognition}
\label{sec:prap}



Goal recognition is the task of recognizing the intended goal that an agent (software or human) aims to achieve from observations of its acting in an environment~\cite{ActivityIntentPlanRecogition_Book2014}.
Observations can be either a sequence of actions performed by the agent or the consequences of such actions, more specifically, properties as logical facts (\exemp, at home, resting).
Furthermore, observations can be either seen as a full sequence of actions or a partial subsequence of actions performed by an agent in an environment.
Plan recognition is a related task to goal recognition, however, the objective of this task is recognizing the plan (\idest, sequence of actions) that an observed agent is executing to achieve a particular goal~\cite{ActivityIntentPlanRecogition_Book2014}. 
Goal and plan recognition in real-world data assume an underlying processing step that translates raw sensor data into some kind of symbolic representation~\cite{ActivityIntentPlanRecogition_Book2014}, as well as a model of the observed agent's behavior generation mechanism.

We use planning domain theories to formalize agents' behavior and the environment description, following the STRIPS formalism proposed by~\citeauthor{Fikes1971}~\shortcite{Fikes1971}. A domain model is a tuple $\mathcal{D} = \langle \mathcal{R}, \mathcal{O} \rangle$, where: $\mathcal{R}$ is a set of predicates with typed variables. 
Predicates can be associated to objects in a concrete problem (\idest, grounded) representing logical values. 
Grounded predicates represent logical values according to some interpretation as facts, which are divided into two types: positive and negated facts, as well as constants for truth ($\top$) and falsehood ($\bot$). 
The set $\mathcal{F}$ of positive facts induces the state-space of a planning problem, which consists of the power set $\mathbb{P}(\mathcal{F})$ of such facts, and the representation of individual states  $S \in \mathbb{P}(\mathcal{F})$. 
$\mathcal{O}$ is a set of operators $op = \langle \mathit{pre}(op), \mathit{eff}(op) \rangle$, where $\mathit{eff}(op)$ can be divided into positive effects $\mathit{eff}^{+}(op)$ (add list) and negative effects $\mathit{eff}^{-}(op)$ (delete list).
An operator $op$ with all variables bound is called an action, and the collection of all actions instantiated for a specific problem induces a state transition function $\gamma(S,a)\mapsto \mathbb{P}(\mathcal{F})$ that generates a new state from the application of an action to the current state.
An instantiated action $a$ from an operator $op$ is applicable to a state $S$ iff $S \models \mathit{pre}(a)$ and results in a new state $S'$ such that $S' \gets (S \cup \mathit{eff}^{+}(a))/\mathit{eff}^{-}(a)$.

A planning problem within $\mathcal{D}$ and a set of typed objects $Z$ is defined as $\mathcal{P} = \langle \mathcal{F}, \mathcal{A}, \mathcal{I}, G \rangle$, where: $\mathcal{F}$ is a set of facts (instantiated predicates from $\mathcal{R}$ and $Z$); $\mathcal{A}$ is a set of instantiated actions from $\mathcal{O}$ and $Z$; $\mathcal{I}$ is the initial state ($\mathcal{I} \subseteq \mathcal{F}$); and $G$ is a partially specified goal state, which represents a desired state to be achieved.
A plan $\pi$ for a planning problem $\mathcal{P}$ is a sequence of actions $\langle a_1, a_2, ..., a_n \rangle$ that modifies the initial state $\mathcal{I}$ into a state $S\models G$ in which the goal state $G$ holds by the successive execution of actions in a plan $\pi$. 
Most automated planners use the \emph{Planning Domain Definition Language} (PDDL) as a standardized domain and problem representation medium~\cite{Fox2003}, which encodes the formalism described thus far.

We follow the definition from \citeauthor{Ramirez:2009wo}~\shortcite{Ramirez:2009wo,Ramirez:2010vv} to formalize the problem of goal recognition problem as planning. A goal recognition problem as planning is a tuple $\mathcal{P}_{GR} = \langle \mathcal{D}, \mathcal{F}, \mathcal{I}, \mathcal{G}, O \rangle$, where $\mathcal{D}$ is a planning domain; $\mathcal{F}$ is the set of facts; $\mathcal{I} \subseteq \mathcal{F}$ is an initial state; $\mathcal{G}$ is the set of possible goals, which include a correct hidden goal $G^{*}$ (\idest, $G^{*} \in \mathcal{G}$); and $O = \langle o_1, o_2, ..., o_n\rangle$ is an observation sequence of executed actions, with each observation $o_i \in \mathcal{A}$, and the corresponding action being part of a valid plan $\pi$ that sequentially transforms $\mathcal{I}$ into $G^{*}$. 
The solution for a goal recognition problem is the correct hidden goal $G^{*} \in \mathcal{G}$ that the observation sequence $O$ of a plan execution achieves. 
An observation sequence $O$ contains actions that represent an optimal or sub-optimal plan that achieves a correct hidden goal, and this observation sequence can be full or partial. 
A full observation sequence represents the whole plan that achieves the hidden goal, \idest, 100\% of the actions having been observed. 
A partial observation sequence represents a subsequence of the plan for the hidden goal, such that a certain percentage of the actions actually executed to achieve $G^{*}$ could not be executed. 


\subsection{Planning in Latent Space}
\label{sec:autoencoding}

Most planning algorithms are based on the \textit{factored} transition function $\gamma(S,a)$ that represents states as discrete facts. 
This transition function is usually encoded manually by a domain expert, and virtually all existing goal and plan recognition approaches require varying degrees of domain knowledge in order to recognize from observations. 
Automatically generating of such domain knowledge involves at least two processes: (1) converting real-world data into a factored representation (\idest, the predicates in $\mathcal{R}$); and (2) generating a transition function (\idest, the set of operators $\mathcal{O}$) from traces of the factored representation. 
Although a few approaches have tackled the challenge of applying learning to models of transition functions~\cite{Jimenez:2012wl}, almost no approaches have addressed the problem of generating domain models from real world data. 
Recently, \citeauthor{Asai_AAAI18}~\shortcite{Asai_AAAI18} developed an approach to planning that generates domain models from images of the visualization of the state of simple games and problems, such as the sliding blocks puzzle or towers of Hanoi. 
This approach uses an autoencoder~\cite{Vincent:2008jr} neural network to automatically generate two functions with regard to an input image $\mathcal{X}$ and a latent representation $\mathcal{L}$: an encoder $\phi: \mathcal{X} \mapsto \mathcal{L}$ and a decoder $\psi: \mathcal{L} \mapsto \mathcal{X}$.
In this specific case, the input is a d-dimensional image $\mathbb{R}^{d}$ and the output is an $n \times m$ matrix $\mathbb{R}^{n\times m}$ representing $n$ categorical variables each of which with $m$ categories. 
When $m$ is two, the output of this auto-encoder corresponds to binary variables that can be interpreted as propositional logic symbols comprising the $\mathcal{F}$ component of a planning domain (without the intermediary step of the generating the set $\mathcal{R}$ of predicates). 

The resulting representation in latent space is amenable to automatically inducing a transition function $\gamma$ from pairs of states under the assumption that state transitions correspond exactly to pairs of consecutive images in the observed traces. 
Under this assumption, they generate a large number of propositional actions representing changes between these images as add and delete effects of STRIPS-style actions. 
The resulting domain representation encodes in latent-space the propositional features from the images.  
LatPlan{$\alpha$} is a heuristic-based forward-search planner~\cite{Asai_AAAI18} that uses this representation to plan solutions for problems derived from images of the initial and target state using the encoded domains.
Preliminary experimentation with LatPlan{$\alpha$}~\cite{Asai_AAAI18} shows that heuristics from the planning literature~\cite[Chapter 3]{Geffner:2013hx} are still applicable, however, given the propositional nature of the encoding, they are not so informative. 
Such lack of informativeness provides a challenge to the application of goal and plan recognition approaches in latent-space.
As we see in Section~\ref{sec:goalRecognitionInLatentSpace}, in order to successfully employ efficient goal recognition approaches, we need not only to learn a consistent latent representation of states, but also to use the propositional transition function induced from state pairs to generate STRIPS-style operators. 

\subsection{Goal Recognition in Latent Space}
\begin{figure}[tb!]
    \centering
    \includegraphics[width=0.48\textwidth]{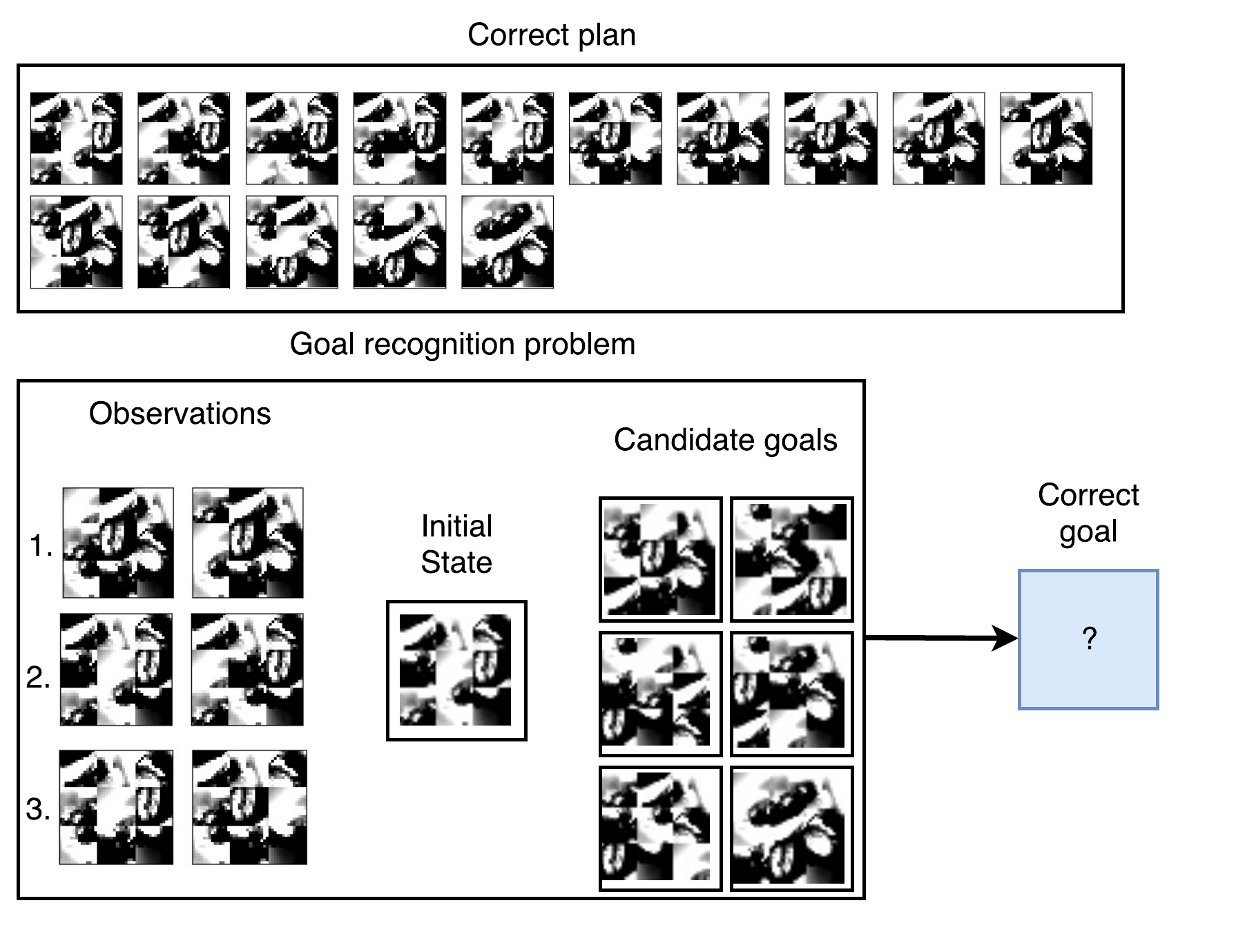}
    \caption{Image of a goal recognition problem.}
    \label{fig:problem_rep}
\end{figure}

Goal recognition in Latent Space is a technique to apply classical goal recognition algorithms in raw data (such as images) by converting it into a latent representation~\cite{amado2018}.
In Figure~\ref{fig:problem_rep}, we provide an example of the goal recognition problem in image domains.
We want to infer what is the correct image configuration that the agent is trying to achieve from the set of candidate goals using only observations consisting of intermediate image configurations. 
As we can see, inferring the correct goal in such task is not trivial, as the small number of observations provide little information.

\label{sec:goalRecognitionInLatentSpace}

To recognize goals in image based domains, \citeauthor{amado2018}~\shortcite{amado2018} proposed four steps. 
First, we must develop an autoencoder capable of creating a latent representation to a state of such image domain. 
Second, since classical goal recognition approaches require a PDDL domain, we need a technique capable of extracting a PDDL domain from the latent representation of the transition of the  domain. 
Third, we must convert to a latent representation a set of images representing, the initial state $\mathcal{I}$, the set of facts $\mathcal{F}$ and a set of possible goals $\mathcal{G}$, where the hidden goal $G^{*}$ is included.
Finally, we can apply goal recognition techniques using the computed tuple $\langle \mathcal{D}, \mathcal{F}, \mathcal{I}, \mathcal{G}, O \rangle$

The encoded representation can be achieved by using the an autoencoder similar to the one described by \citeauthor{Asai_AAAI18}~\shortcite{Asai_AAAI18} with the Gumbel softmax \cite{gumbel1954statistical} activation function.
Each domain requires one autoencoder capable to converting an image state of the domain to a latent representation.
We preprocessed these images by applying a grayscale filter and then binarizing the resulting image.
With a trained autoencoder for each domain it is possible to output a PDDL domain using the Action Learner develop by \citeauthor{amado2018}~\shortcite{amado2018}. This PDDL domain will have a compressed number of actions to improve the speed of the goal recognition process.

\begin{figure}[tb!]
    \centering
    \includegraphics[width=0.40\textwidth]{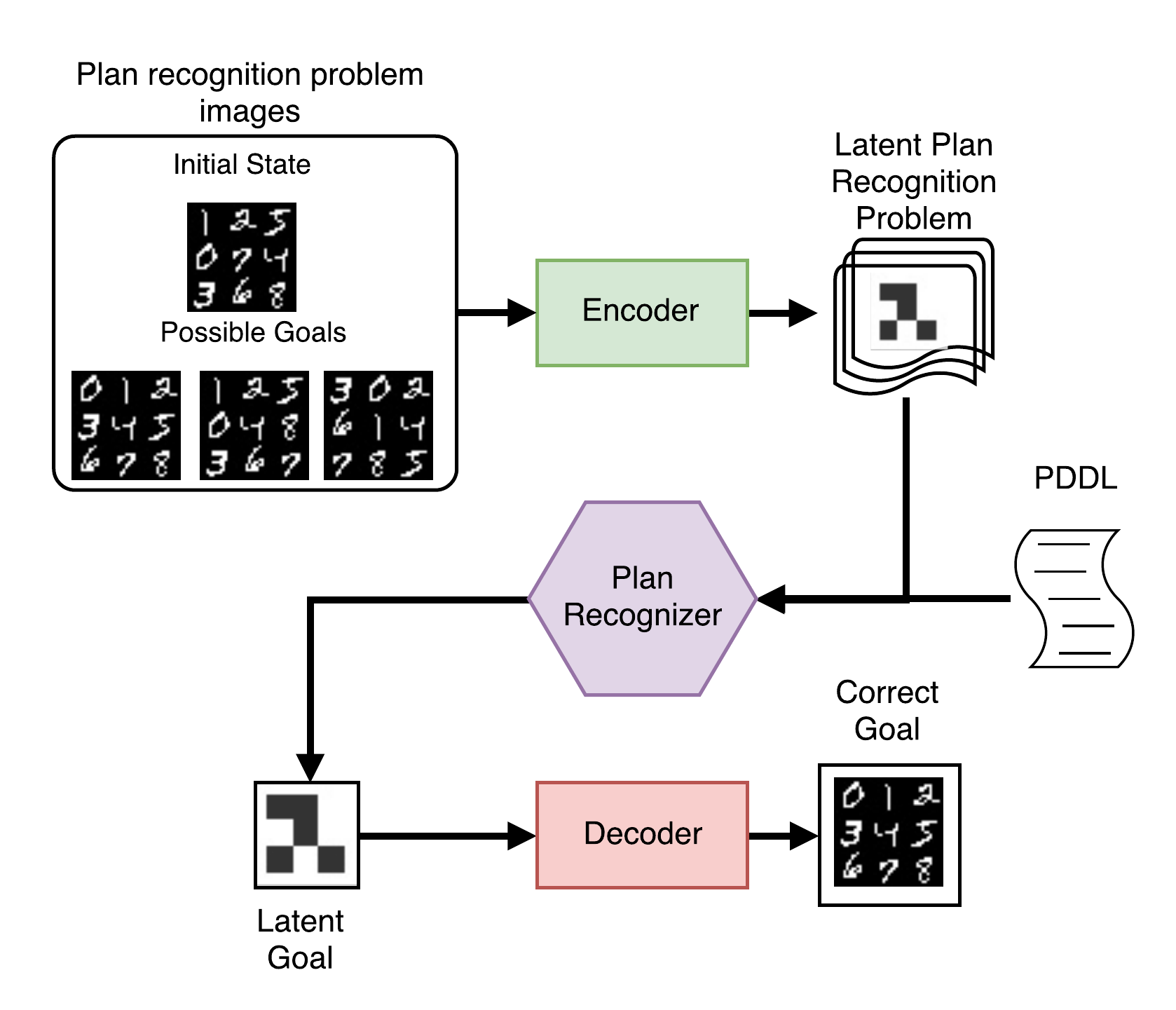}
    \caption{Image goal recognition process.}
    \label{fig:recognition}
\end{figure}

Following Section~\ref{sec:prap}, we represent a goal recognition problem by the tuple $\mathcal{P}_{GR} = \langle \mathcal{D}, \mathcal{F}, \mathcal{I}, \mathcal{G}, O \rangle$. 
We extract the domain $\mathcal{D}$ using the Action Leaner, and the facts $\mathcal{F}$ represented by the latent space representation. 
We compute the initial state $\mathcal{I}$, a set of candidate goals $\mathcal{G}$, and finally a set of observations $O$. 
To compute $\mathcal{I}$ and the set of goals $\mathcal{G}$, we use the image representations of these states and convert them to latent representation using the trained autoencoder. 
To derive the observations $O$, we take pairs of images representing of the environment. 
These images are encoded to the latent representation, and then by using the PDDL domain we extracted, we compute which action from the PDDL domain was responsible for such state transition. 
%
After building a goal recognition problem, we can now apply off-the-shelf goal recognition techniques, such as~\cite{Ramirez:2009wo,Ramirez:2010vv,Sohrabi:2016uw,PereiraOrenMeneguzzi2017}. 
The output of such techniques is the goal with highest probability of being the correct one, in the latent space representation. 
We then decode the inferred goal, obtaining its image representation using the decoder. 
This process is illustrated in Figure~\ref{fig:recognition}.

\subsection{Long Short-Term Memory Networks}
\label{sec:lstm}

A Recurrent Neural Network (RNN) is a network that attempts to model a sequence of dependent events occurring through time such as financial time series \cite{AkitaEtAl2016}, language modeling \cite{SundermeyerEtAl2015} and so on. 
The recurrence is performed by feeding the input layer of the network at time $t+1$ with the output of the network layer at time $t$, keeping a ``memory'' of the past events. 
Unfortunately, RNNs suffer with well-known vanishing gradient problem \cite{BengioEtAl1994}, \idest, the gradients that are backpropagated thought the network during the training phase tend to decay or grow exponentially. 
Therefore, as dependencies in RNNs get longer, the gradient calculation becomes unstable, limiting the network to learn long-range dependencies.

\begin{figure}[tb!]
    \centering
    \includegraphics[width=0.48\textwidth]{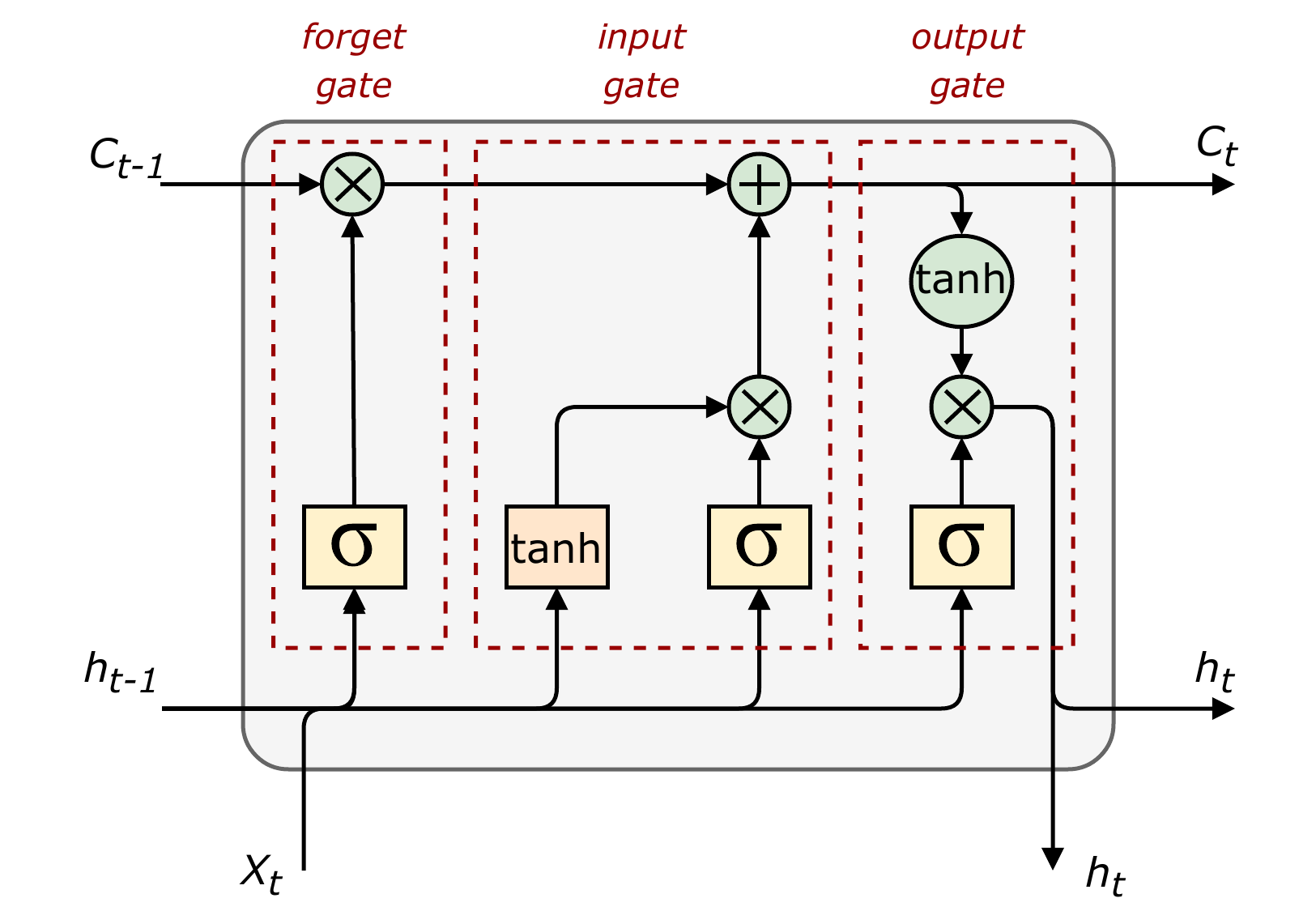}
    \caption{Internal structure of the Long-Short Term Memory cell.}
    \label{fig:lstmcell}
\end{figure}

In order to get rid of the vanishing gradient problem, \citeauthor{Hochreiter:1997:LSM:1246443.1246450}~\shortcite{Hochreiter:1997:LSM:1246443.1246450} propose an RNN architecture called Long Short-Term Memory (LSTM) network that modifies the original recurrent cell such that vanishing and exploding gradients are avoided, whereas the training algorithm is left unchanged. 
An LSTM cell contains mainly four components called cell state, forget gate, input gate and output gate. 
The cell state ($C$) is responsible for passing the information through the cell to the next LSTM cell, while being changed by the gates. 
The forget gate decides what information should be forget from the previous cell state. This gate contains a \emph{sigmoid} ($\sigma$) layer that outputs a number between 0 and 1, where 1 means ``keep all information'' and 0 means ``get rid of this information''. 
Input gate decides what information should be stored in the cell state by applying a \emph{sigmoid} layer to decide what information to keep and a hyperbolic tangent (\emph{tanh}) layer to select new candidates to the cell state, performing an update to the cell state. 
Finally, the output gate decides what information should be propagated forward by performing a pointwise multiplication of a \emph{sigmoid} layer, which decides what part of the input should be forwarded, and a the cell state filtered by a \emph{tanh} operation. 
Figure \ref{fig:lstmcell} illustrates the LSTM cell with its respective gates, where yellow boxes represent layers, elements in green represent pointwise operations ($\otimes$ pointwise multiplication, $\oplus$ pointwise addition and \emph{tanh} pointwise hyperbolic tangent function), merging arrows represent the concatenation of elements and forking arrows represent the copy of the content to multiple points. 

Therefore, an LSTM performs a classification problem by receiving a streamline of ordered data as input and returning a class based on the data sequence received. 
In this work, we use plan traces as input sequences and their corresponding goals as training class when training the network. 
Hence, the network learns the agent's goal based on the sequence of actions performed by the agent. 

\section{Goal Recognition in Latent Space using LSTM}
Current approaches to recognize goals in latent space require enough data to build a complete PDDL domain~\cite{amado2018}. To avoid the need of such high amount of domain knowledge, we propose the usage of a machine learning model capable of recognizing goals using only plan traces as training data. 

\begin{figure}[h!]
    \centering
    \includegraphics[width=0.50\textwidth]{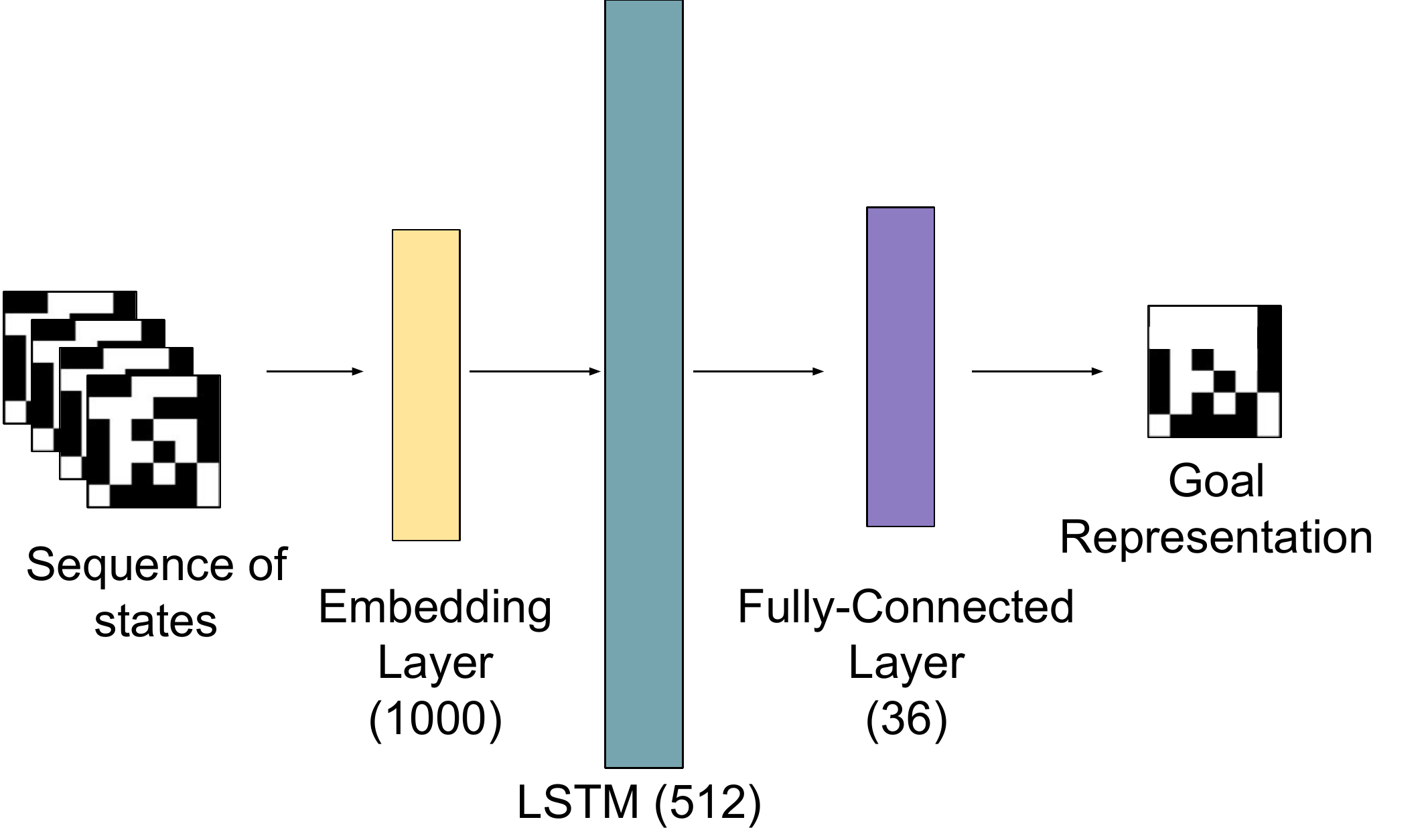}
    \caption{LSTM Architecture}
    \label{fig:lstm_structure}
\end{figure}

Our LSTM consists of three main layers.
First, we use an embedding layer to convert our input sequence into a dense representation with a dimension of 1000 that will feed the LSTM units.
Second, we use an LSTM layer containing 512 units.
Finally, a fully connected layer receives the output from LSTM and generates the goal representation with 36 output neurons.
We use sigmoid activation on the neurons from the output layer and a binary cross entropy loss using RMSprop as optimizer.
Figure~\ref{fig:lstm_structure} illustrates our LSTM architecture.

\begin{figure}[b!]
    \centering
    \includegraphics[width=0.5\textwidth]{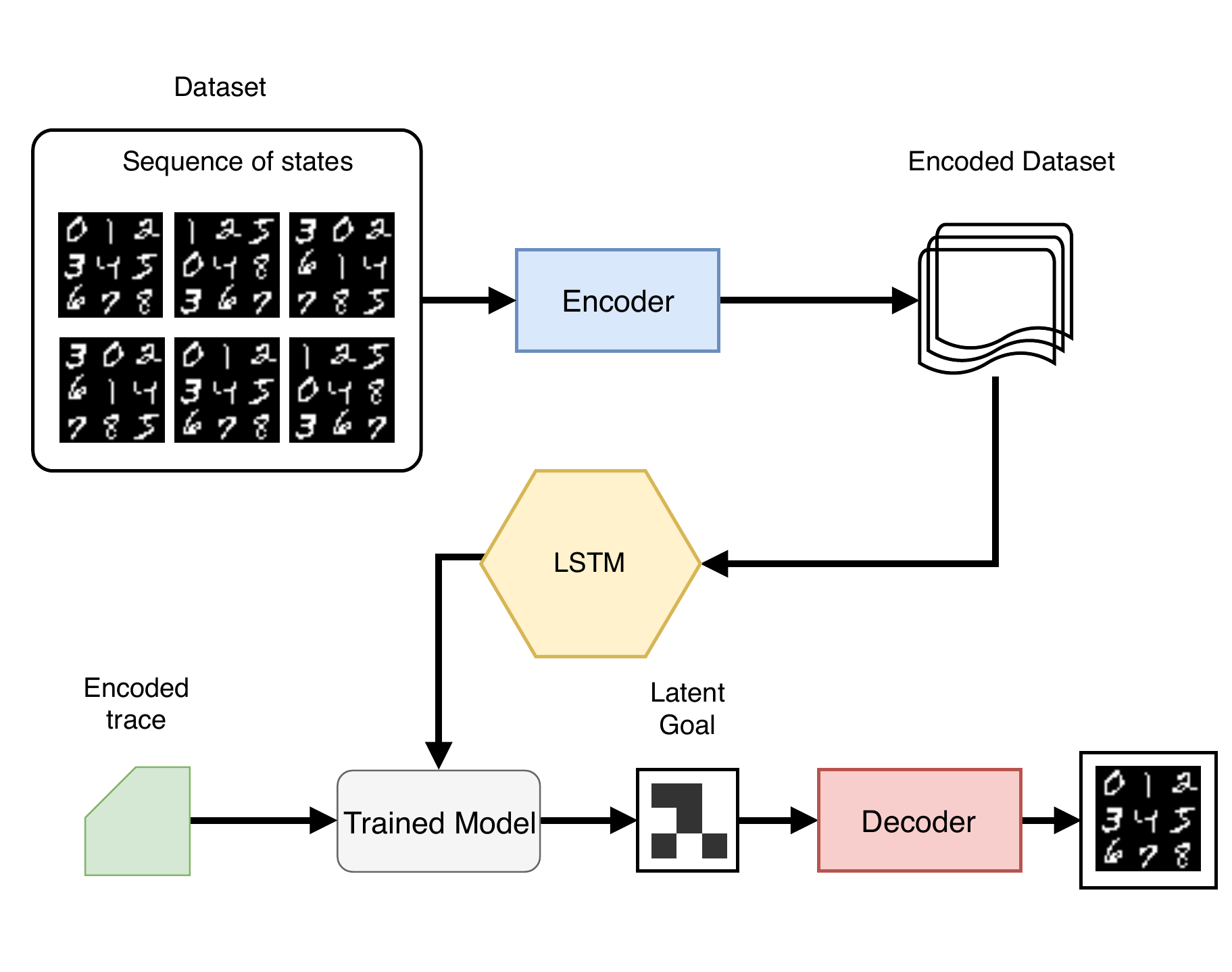}
    \caption{Goal recognition using LSTMs}
    \label{fig:lstm}
\end{figure}

In order to create a model to recognize goals, we train an LSTM that receives a sequence of encoded states and predicts an encoded goal.
To perform a fair comparison to the state-of-the-art, we use as input encoded states generated by the encoder module from the autoencoder created by \citeauthor{Asai_AAAI18}~\shortcite{Asai_AAAI18}.
Thus, we convert each image-state into a latent representation (a 6x6 binary matrix).
Figure~\ref{fig:lstm} illustrates the process of training and testing our LSTM model, we highlight three main steps of such process.
First, given a set of image-states representing a sequence of states and the goal of a certain plan, we use the encoder to generate the latent representation for each image.
Second, using the representations, we train the LSTM to predict the goal given the states.
The output is a representation of this goal.
Finally, we use the decoder from \citeauthor{Asai_AAAI18} autoencoder to convert the produced representation into an image.

To train the LSTM network, we require data extracted from plans for each domain.
We use plan traces generated by \citeauthor{amado2018}~\shortcite{amado2018}, observing the states that were reached in each plan. 
Each trace generated a list of states, and then we included the goal of each trace as a class to the LSTM.
To improve accuracy in low observability scenarios, we included partially observable traces (which means some states were removed from the plan trace), including 10\%, 30\%, 50\%, 70\% of observability. 
During the training phase, we use early stopping to avoid overfitting and set a limit of 10,000 epochs.
Early stopping monitors validation loss ensuring training will stop when loss stops decreasing.

We manipulate LSTM inputs by converting the latent representations into a specific encoding.
In our specific case, we turn each state into an integer number, thus, we differentiate them simplifying the input.
An entry example of such model is: \textit{22, 23, 33, 48, 12}, where each number is a specific state from the state-space in its domain and the sequence is an entire plan.
The output layer is 36 binary neurons, which we use to rebuild the latent representation by reshaping it into a 6x6 matrix.

\section{Experiments}
\label{sec:experiments}

\begin{table*}[t!]
\fontsize{6}{5}\selectfont
\centering
\caption{Experimental results on Goal Recognition in Latent Space.}
\def\arraystretch{0.55}
\begin{tabular}{|c|c|cc|ccc|ccc|ccc|}
\hline
\multicolumn{4}{|c|}{}    & \multicolumn{3}{c|}{\textsc{POM} ($\mathit{h_{gc}}$)} & \multicolumn{3}{c|}{\textsc{LSTM} } & \multicolumn{3}{c|}{\textsc{RG}}  \\ \hline
\textbf{Domain} & $|\mathcal{G}|$    & (\%) \textbf{Obs} & $|O|$   & \begin{tabular}[c]{@{}c@{}}\textbf{Time (s))}\\$\theta$ (0 / 10)\end{tabular}  & \begin{tabular}[c]{@{}c@{}}\textbf{Accuracy} \%\\$\theta$ (0 / 10)\end{tabular} & \begin{tabular}[c]{@{}c@{}} \textbf{Spread in} $\mathcal{G}$\\$\theta$ (0 / 10)\end{tabular} & \begin{tabular}[c]{@{}c@{}}\textbf{Time (s)}\\ \end{tabular}      & \begin{tabular}[c]{@{}c@{}}\textbf{Accuracy} \%\\\end{tabular} & \begin{tabular}[c]{@{}c@{}} \textbf{Spread in} $\mathcal{G}$\\ \end{tabular}  & \textbf{Time (s)} & \textbf{Accuracy} \% & \textbf{Spread in} $\mathcal{G}$ \\ \hline
&      
& 10  & 1.2 
	&  0.591 / 0.603  &  33.3\% / 83.3\% &  1.6 / 4.0
    &  0.346  &  16.6\%  &  1.0
    &  21.25  &  100.0\% &  6.0       \\ &      
& 30  & 3.0 
	&  0.612 / 0.625  &  33.3\% / 83.3\%  &  1.4 / 2.8
    &  0.335 &   100.9\%  &  1.0
    &  22.26  &  100.0\% &  4.8        \\
MNIST  & 6.0 
& 50  & 4.0 
	&  0.673 / 0.677  &  60.0\% / 100.0\% &  2.2 / 3.0
    &  0.326 &  100.0\%   &  1.0
    &  22.48  &  100.0\% &  4.8        \\ &      
& 70  & 5.8 
	&  0.698 / 0.703  & 100.0\% / 100.0\% &  2.4 / 3.0      
    &  0.394  &  100.0\%  &  1.0           
    &  23.53  &  100.0\% &  3.2       \\ &      
& 100 & 7.8  
	&  0.724 / 0.730 &  100.0\% / 100.0\%  &  2.4 / 3.0       
    &  0.357  &  100.0\%  &  1.0
    &  26.34  &  100.0\% &  3.4      \\ \hline
&      
& 10  & 1.8 
	&  0.013 / 0.014  &  16.6\% / 83.3\%  &  1.0 / 3.8
    &  0.335  &  50\%  &  1.0
    &  1.02  &  83.3\% &  5.6       \\ &
& 30  & 4.8 
	&  0.015 / 0.017  &  16.6\% / 100.0\%  &  1.0 / 4.8
    &  0.366 &  100.0\% &  1.0
    &  1.38  &  83.3\% &  3.8        \\
Mandrill  & 6.0 
& 50  & 6.0 
	&  0.018 / 0.018  &  33.3\% / 83.3\%  &  1.1 / 4.8
    &  0.389  &  100.0\%  &  1.0
    &  1.44  &  83.3\% &  4.1        \\ &      
& 70  & 8.1 
	&  0.020 / 0.021  &  50.0\% / 83.3\%  &  1.3 / 4.3
    &  0.353  &  100.0\% &  1.0
    &  1.68  &  66.6\% &  1.8       \\ &      
& 100 & 11.3  
	&  0.022 / 0.023  &  66.6\% / 100.0\% &  1.8 / 5.16
    &  0.347 &  100.0\%  &  1.0
    &  1.71  &  66.6\% &  1.8      \\ \hline
&      
& 10  & 1.5 
	&  0.166 / 0.178  &  33.3\% / 66.6\%  &  2.3 / 4.8
    &  0.375  &  83.3\%  &  1.0
    &  1.35  &  83.3\% &  4.1       \\ &
& 30  & 4.0 
	&  0.181 / 0.190  &  66.6\% / 66.6\%  &  4.1 / 5.1
    &   0.423  &  83.3\%  &  1.0
    &  1.57  &  83.3\% &  3.0        \\
Spider  & 6.0 
& 50  & 5.6 
	&  0.193 / 0.199  &  50.0\% / 83.3\%  &  3.5 / 5.5
    &  0.431  &  100.0\%  &  1.0
    &  1.66  &  83.3\% &  2.8        \\ &      
& 70  & 7.5 
	&  0.201 / 0.205  &  83.3\% / 83.3\% &  4.6 / 5.5
    &  0.384  &  100.0\%  &  1.0
    &  1.79  &  66.6\% &  2.3       \\ &      
& 100 & 10.5  
	&  0.208 / 0.217  &  100.0\% / 100.0\% &  5.5 / 6.0
    &  0.368  &  100.0\%  &  1.0
    &  2.04  &  66.6\% &  1.1      \\ \hline
&      
& 10  & 1.0 
	&  0.831 / 0.902  &  33.3\% / 33.3\% &  1.5 / 3.0
    &  0.315  &  83.3\%  &  1.0
    &  42.52  &  100.0\% &  6.0       \\ &
& 30  & 1.6 
	&  0.884 / 1.09  &  33.3\% / 66.6\%  &  1.5 / 4.3
    &  0.317  &  100.0\%  &  1.0
    &  43.07  &  100.0\% &  5.5        \\
LO Digital  & 6.0 
& 50  & 2.5 
	&  0.915 / 1.13  &  33.3\% / 83.3\%  &  1.5 / 4.5
    &  0.336  &  100.0\%  &  1.0
    &  43.41  &  83.3\% &  5.1        \\ &      
& 70  & 3.6 
	&  0.970 / 1.19  &  83.3\% / 100.0\% &  3.6 / 4.5
    &  0.371  &  83.3\%  &  1.0
    &  43.78  &  100.0\% &  4.8       \\ &      
& 100 & 4.3  
	&  1.12 / 1.24  &  100.0\% / 100.0\% &  2.6 / 4.3
    &  0.330  &  83.3\%  &  1.0
    &  43.91  &  100.0\% &  4.8      \\ \hline 
&      
& 10  & 1.0 
	&  1.16 / 1.21  &  16.6\% / 16.6\% &  1.0 / 3.0
    &  0.376  &  66.6\%  &  1.0
    &  121.97  &  100.0\% &  5.8       \\ &
& 30  & 1.6 
	&  1.25 / 1.39  &  16.6\% / 50.0\%  &  1.0 / 3.8
    &  0.320  &  100.0\%  &  1.0
    &  123.92  &  100.0\% &  5.0        \\
LO Twisted  & 6.0 
& 50  & 2.1 
	&  1.33 / 1.46  &  16.6\% / 50.0\%  &  1.0 / 4.5
    &  0.339  &  100.0\%  &  1.0
    &  124.42  &  100.0\% &  5.6        \\ &      
& 70  & 3.3 
	&  1.48 / 1.50  &  16.6\% / 83.3\% &  1.0 / 3.3
    &  0.312  &  100.0\%  &  1.0
    &  127.22  &  100.0\% &  5.5       \\ &      
& 100 & 4.3  
	&  1.57 / 1.62  &  100.0\% / 100.0\% &  2.3 / 5.0
    &  0.327  &  100.0\%  &  1.0
    &  129.99  &  100.0\% &  5.5      \\ \hline
&      
& 10  & 1.0 
	&  0.304 / 0.318  &  33.3\% / 66.6\% &  1.0 / 2.3
    &  0.334  &  66.6\%  &  1.0
    &  6.08  &  100.0\% &  4.0       \\ &
& 30  & 3.0 
	&  0.316 / 0.320  &  100.0\% / 100.0\%  &  4.0 / 4.0
    &  0.365  &  100.0\%  &  1.0
    &  6.21  &  100.0\% &  4.0        \\
Hanoi  & 4.0 
& 50  & 4.3 
	&  0.322 / 0.337  &  100.0\% / 100.0\%  &  4.0 / 4.0
    &  0.371  &  100.0\%  &  1.0
    &  7.01  &  66.6\% &  3.3        \\ &      
& 70  & 6.0 
	&  0.345 / 0.354  &  100.0\% / 100.0\% &  4.0 / 4.0
    &  0.372  &  66.6\%  &  1.0
    &  7.26  &  100.0\% &  4.0       \\ &      
& 100 & 8.3  
	&  0.354 / 0.362  &  100.0\% / 100.0\% &  4.0 / 4.0
    &  0.329  &  66.6\%  &  1.0
    &  8.19  &  100.0\% &  4.0      \\ \hline    
\end{tabular}
\label{tab:GoalRecognitionResults}
\end{table*}

\begin{table*}[t!]
\fontsize{6}{5}\selectfont
\centering
\def\arraystretch{0.55}
\caption{Experimental results on Goal Recognition using handmade domains.}
\begin{tabular}{|c|c|cc|ccc|ccc|}
\hline
\multicolumn{4}{|c|}{}    & \multicolumn{3}{c|}{\textsc{POM} ($\mathit{h_{gc}}$)} & \multicolumn{3}{c|}{\textsc{RG}}  \\ \hline
\textbf{Domain} & $|\mathcal{G}|$    & (\%) \textbf{Obs} & $|O|$   

& \begin{tabular}[c]{@{}c@{}}\textbf{Time (s))}\\$\theta$ (0 / 10)\end{tabular}  & \begin{tabular}[c]{@{}c@{}}\textbf{Accuracy} \%\\$\theta$ (0 / 10)\end{tabular} & \begin{tabular}[c]{@{}c@{}} \textbf{Spread in} $\mathcal{G}$\\$\theta$ (0 / 10)\end{tabular} 

& \textbf{Time (s)} & \textbf{Accuracy} \% & \textbf{Spread in} $\mathcal{G}$ \\ \hline
&      
& 10  & 1.6
	&  0.010 / 0.012  &  66.6\% / 100.0\% &  1.6 / 2.3
    &  0.075  &  33.3\% &  1.3       \\ &      
& 30  & 4.0 
	&  0.011 / 0.012  &  66.6\% / 100.0\%  &  1.0 / 1.3
    &  0.080  &  100.0\% &  2.3        \\
Hanoi  & 4.0 
& 50  & 6.3 
	&  0.012 / 0.013  &  66.6\% / 100.0\% &  1.0 / 1.6
    &  0.085  &  100.0\% &  1.3        \\ &      
& 70  & 8.6 
	&  0.013 / 0.013  & 100.0\% / 100.0\% &  1.3 / 1.3      
    &  0.091  &  100.0\% &  1.3       \\ &      
& 100 & 11.6
	&  0.013 / 0.013 &  100.0\% / 100.0\%  &  1.6 / 2.0       
    &  0.098  &  100.0\% &  1.3      \\ \hline
&      
& 10  & 1.0 
	&  0.098 / 0.111  &  16.6\% / 33.3\%  &  1.0 / 2.6
    &  0.179  &  100.0\% &  4.8       \\ &
& 30  & 3.0 
	&  0.109 / 0.120  &  66.6\% / 100.0\%  &  1.1 / 2.3
    &  0.188  &  100.0\% &  1.3        \\
8-Puzzle  & 6.0 
& 50  & 4.0 
	&  0.117 / 0.129  &  66.6\% / 100.0\%  &  1.0 / 2.0
    &  0.191  &  100.0\% &  1.3        \\ &      
& 70  & 5.3 
	&  0.121 / 0.135  &  100.0\% / 100.0\%  &  1.0 / 1.8
    &  0.210  &  100.0\% &  1.0       \\ &      
& 100 & 7.3  
	&  0.133 / 0.141  &  100.0\% / 100.0\% &  1.0 / 1.1
    &  0.246  &  83.3\% &  1.1      \\ \hline
&      
& 10  & 1.0 
	&  0.689 / 0.766  &  33.3\% / 66.6\%  &  1.3 / 3.8
    &  5.76  &  100.0\% &  5.6       \\ &
& 30  & 1.6 
	&  0.721 / 0.780  &  50.0\% / 83.3\%  &  1.6 / 4.5
    &  5.79  &  100.0\% &  5.3        \\
Light-Out  & 6.0 
& 50  & 2.6 
	&  0.788 / 0.811  &  33.3\% / 100.0\%  &  2.6 / 5.3
    &  5.82  &  100.0\% &  5.4        \\ &      
& 70  & 3.6 
	&  0.804 / 0.849  &  66.6\% / 100.0\% &  3.8 / 5.0
    &  5.90  &  100.0\% &  5.3       \\ &      
& 100 & 4.3  
	&  0.875 / 0.956  &  100.0\% / 100.0\% &  4.6 / 6.0
    &  5.93  &  100.0\% &  4.8      \\ \hline
\end{tabular}
\label{tab:GoalRecognitionResults_Handmade}
\end{table*}


\subsection{Datasets}
\label{sec:datasets}

In order to evaluate our LSTM approach, we generated a number of image-based datasets based on existing goal recognition problems~\cite{ramon_fraga_pereira_2017_825878,Asai_AAAI18,amado2018}. 
We have two main experimental objectives: first, we want to compare the performance of our LSTM approach with the existing approaches to goal recognition in latent space, using problems were the goal of each problem is contained in the dataset; second, we want to evaluate the performance of our LSTM approach when dealing with problems were the goal is not contained as a class in our dataset.
Our evaluation dataset thus are two distinct datasets. The first, a dataset containing the exact problems used in \cite{amado2018} to evaluate their latent goal recognition approach. The second, a dataset containing new goal recognition problems, with distinct goals where these goals do not appear as a class in the training dataset for the LSTM. 
In order to generate such traces, we use a standard PDDL planner~\cite{Helmert06thefast} to search for a plan for a set of randomly generated goals.
From the resulting traces, we can generate the observations at various levels of observability by omitting the states resulting from a percentage of the actions generated by the planner. 

\begin{figure}
\begin{subfigure}{.15\textwidth}
  \centering
  \includegraphics[scale=1.0]{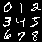}
  \caption{MNIST}
  \label{fig:sfig1}
\end{subfigure}%
\begin{subfigure}{.15\textwidth}
  \centering
  \includegraphics[scale=1.0]{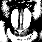}
  \caption{Mandrill}
  \label{fig:sfig2}
\end{subfigure}%
\begin{subfigure}{.15\textwidth}
  \centering
  \includegraphics[scale=1.0]{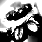}
  \caption{Spider}
  \label{fig:sfig3}
\end{subfigure}
\vspace{1cm}
\begin{subfigure}{.15\textwidth}
  \centering
  \includegraphics[scale=1.0]{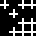}
  \caption{LO Digital}
  \label{fig:sfig4}
\end{subfigure}%
\begin{subfigure}{.15\textwidth}
  \centering
  \includegraphics[scale=1.0]{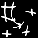}
  \caption{LO Twisted}
  \label{fig:sfig5}
\end{subfigure}%
\begin{subfigure}{.15\textwidth}
  \centering
  \includegraphics[scale=1.0]{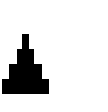}
  \caption{Hanoi}
  \label{fig:sfig6}
\end{subfigure}
\vspace{-2em}
\caption{Sample state for each domain.}
\label{fig:fig}
\end{figure}

Using this method to produce experimental datasets, we generated PDDL domains and images for six different datasets:
\begin{itemize}
\item three variations of the 8-Puzzle, whose goal to order a set of pieces when you can only move the blank space:
\begin{itemize}
\item the MNIST 8-puzzle uses the handwritten digits from the MNIST dataset as the pieces of the puzzle, with the number 0 representing the blank space, as illustrated in Figure~\ref{fig:sfig1}---every image of the dataset uses the same handwritten digit for every repeating number;
\item the Mandrill 8-puzzle uses the image of a Mandrill, shown in Figure~\ref{fig:sfig2}---we use the mandrill's right eye as the blank space;
\item the Spider 8-puzzle uses the image of a Spider, shown in Figure~\ref{fig:sfig3}---like the mandrill data set, we use the spider's right eye as the blank space;
\end{itemize}
\item two variations of the Lights-out puzzle game \cite{Fleischer2013}, which consists of a 4 by 4 grid of lights that can be turned on and off, and which starts with a random number of lights initially on---toggling any of the lights also toggles every adjacent light---the objective is to turn every light off;
\begin{itemize}
\item lights-out digital (LO Digital) is a standard lights out representation using crosses to represent when a light is on, illustrated in Figure~\ref{fig:sfig4}; 
\item lights-out twisted (LO Twisted) is a variation of the digital version of lights out such that the image representation undergoes a distortion filter, twisting the exact position of each light, as seen in Figure~\ref{fig:sfig5}; and
\end{itemize}
\item the Tower of Hanoi puzzle, which consists of stacked disks of different sizes and stakes---the objective is to move every disk to a different stack, and we we use a version of the puzzle with three disks and four stakes illustrated in Figure~\ref{fig:sfig6}. 
\end{itemize}
Table~\ref{tab:dataset_spec} describes our dataset specifications, such as domains, number of traces for each domain, and training time.
As we can see, most domains have more than 1000 traces with training times smaller than 5 minutes.
As outliers, LO digital and twisted have small number of traces, it occurs because their plans are relatively smaller when compared to the other domains, which limits the variety of traces.

\begin{table}[ht]
	\centering
    \caption{Dataset specifications }
    \begin{tabular}{l|c|c}
        \toprule
    	\textbf{Domain} & \textbf{$\#$ of Traces} & \textbf{Training Time (seconds)} \\
        \midrule\midrule
        Hanoi      & 1552 &  22.57 \\
        LO Digital &  230 & 294.58 \\
        LO Twisted &  224 &  64.33 \\
        Mandrill   & 2520 &  14.87 \\
        MNIST      & 1427 & 211.42 \\
        Spider     & 2216 & 333.25 \\
        \bottomrule
	\end{tabular}
	\label{tab:dataset_spec}
\end{table}

\subsection{Goal Recognition}
\label{sec:goalRecognitionExperiments}

To compare our approach with the existing approaches of goal recognition for latent space, we use the exact same dataset used in \cite{amado2018}. 
This dataset consists of 6 distinct problems for each domain, where each problem has at least 4 distinct candidate goals. The candidate goals are not necessary for the LSTM. 
From each of these problems (\idest, the initial states and candidate goals), we generate 5 different conditions for the goal recognition algorithm, by altering the level of observability available to the algorithm. 
We set five different percentages of observability: $100\%$, $70\%$, $50\%$, $30\%$ and $10\%$. 

The observations from the Dataset described in Section~\ref{sec:datasets} are pruned so that only the specified fraction of the original observations are left. 
We use two goal recognition approaches to compare with our  LSTM approach (LSTM in Table~\ref{tab:GoalRecognitionResults}): the landmark-based heuristics $h_{gc}$ (Goal Completion Heuristic) developed by Pereira, Oren, and Meneguzzi (POM in in Table~\ref{tab:GoalRecognitionResults})~in~\cite{PereiraOrenMeneguzzi2017}, and the most accurate approach developed by \citeauthor{Ramirez:2009wo}~\shortcite{Ramirez:2009wo} (RG in in Table~\ref{tab:GoalRecognitionResults}). 
These two approaches are the current state-of-the-art in goal and plan recognition in terms of time and accuracy, respectively. 

Table~\ref{tab:GoalRecognitionResults} summarizes goal recognition performance of each approach using the latent representation and learned PDDL encoding provided in \cite{amado2018}, for all domains in the dataset and three different goal recognition approaches. In the LSTM approach, the learned PDDL is not needed to perform goal recognition, only the encoded traces. In this comparison, every hidden goal was included in the LSTM training set at least once. We guarantee that the traces used in this comparison were not included in the training set.
Each row of this table shows averages for the number of candidate goals $|\mathcal{G}|$; 
the percentage of the plan that is actually observed (\%) Obs; 
the average number of observations per problem $|O|$; 
and, for each goal recognition approach, the time in seconds to recognize the goal given the observations; 
the Accuracy \% with which the approaches correctly infer the hidden goal; 
and Spread in $\mathcal{G}$, representing the average number of returned goals. For the LSTM is always one, as it always returns one goal.
As we can see, the LSTM achieved overall good accuracy across all domains and observability scenarios. The execution time was between 0.3 and 0.5 seconds.  
While the RG approach has a good accuracy, it does so with a large spread and long execution times. 
This trade-off is highlighted in the most complex domains, such as Lights out digital and lights out twisted. 
The POM approach also struggled with high spread in some domains, such as the hanoi domain, but was much faster than RG in all scenarios.
Overall the LSTM achieved better results, considering it returns always one goal and the other approaches struggled with high spread. As we can see, for recognizing goals that are contained in the training set, the LSTM is a promising approach that does well in both speed and accuracy.

For comparison, Table~\ref{tab:GoalRecognitionResults_Handmade} shows the results of solving these problems with hand made PDDL domains. 
Since there is no learning inaccuracies in the PDDL of such domains, the results are often superior than the learned models. 
However, in the lights out model, we can see that the approaches also struggle with a high amount of spread.

\begin{table}[h!]
   \centering
   \caption{LSTM results with unknown goals.}
   \begin{tabular}{l|c|c|c}
      \toprule
      \multirow{2}{*}{\textbf{Domain}} & \textbf{Reconstruction} & \multirow{2}{*}{\# \textbf{Problems}} & \textbf{Correct}     \\
                              & \textbf{Accuracy} (\%)  &                              & \textbf{Predictions} \\
      \midrule\midrule
      MNIST      & 48.6\%  & 30  & 0\%  \\
      Mandrill   & 53.6\%  & 30  & 0\%  \\
      Spider     & 58.4\%  & 30  & 0\%  \\
      LO Digital & 53\%    & 30  & 0\%  \\
      LO Twisted & 51.2\%  & 30  & 0\%  \\
      \bottomrule
	\end{tabular}
	\label{tab:LSTM_UGoals}
\end{table}

In Table \ref{tab:LSTM_UGoals} we display the results when dealing with goals that are not contained in the the training set.
The test set consists of 6 distinct problems with distinct goals, where each problem generates 5 traces using different observability (10, 30, 50, 70, 100\%).
The LSTM was unable to recognize any of the goals that are not contained in the dataset. We present the reconstruction accuracy, that estimates how close was the LSTM to reconstruct the correct goal.
There is no direct comparison to other goal recognition approaches, as there is a training data is used in the other approaches.
As we can see, our approach needs the goal to be contained in the training set, as the LSTM network is unable to reconstruct a goal that it has not seen.
In such scenarios enumerating every possible goal is not recommended, as the number of possible states (and so goals) in a 8-Puzzle problem is 362,880.
Thus our approach by encoding classes for classification is very promising, as long as the training set contains many goals (and thus classes), as it removes the burden of enumerating every classes. 



\section{Related Work}
\label{sec:relatedWork}

\citeauthor{Min2014DeepLG}~\shortcite{Min2014DeepLG} propose a deep LSTM network approach capable of recognizing goals of a player in an educational game scenario. 
The dataset used for training the deep LSTM is a player behavior corpus consisting of distinctive player actions. 
The challenge comes from recognizing goals when handling uncertainty from noise input and non-optimal player behavior. 
The LSTM is able to do standard metric-based goal recognition and online goal recognition, as information is fed.

Although the work is very similar to ours, the main difference is that the entirety of the goals are already known in the work proposed by \citeauthor{Min2014DeepLG}, while in our work, we try to reconstruct the goal from the observation traces. 
This is a significant difference, because our approach tries to recognize goals without any domain knowledge from a domain expert, making our approach completely domain independent.  The results will vary depending how many times the goal appears in the training data.

\citeauthor{Granada2017}~\shortcite{Granada2017} develop a hybrid approach that combines activity and plan recognition for video streams. This approach uses deep learning to analyze video data (frames) in order to identify individual actions in a scene, and based on this set of identified actions, a plan recognition algorithm then uses a plan library describing possible overarching activities for recognizing the ultimate goal of the subject in a video. 

\citeauthor{Asai_AAAI18}~\shortcite{Asai_AAAI18} develop a planning architecture capable of planning using only pairs of images (representing, respectively, the initial and goal states) from the domain by converting the images into a latent space representation.
Their architecture consists of a variational autoencoder (VAE) followed by an off-the-shelf planning algorithm.
The architecture convert images into discrete latent vectors using the VAE, and uses the information in such latent vectors to plan over the images and find a sequence of actions that transforms the state into one matching the goal image.


\citeauthor{amado2018}~\shortcite{amado2018} develop an approach to recognize goals in image domains, by converting the images to a latent representation, deriving a PDDL from domain knowledge converted to a latent representation and then applying off-the-shelf goal recognition algorithms. Our work extends this approach by proposing an LSTM as a goal recognizer. The difference is a trade-off between domain knowledge, since the LSTM does not require a PDDL domain, and training dataset using plan traces that is necessary to train the LSTM network.

\citeauthor{Ramirez:2009wo}~\shortcite{Ramirez:2009wo} propose planning approaches for goal and plan recognition, and instead of using plan-libraries, they model the problem as a planning domain theory with respect to a known set of candidate goals. This work uses a modified heuristic, an optimal and modified sub-optimal planner to determine the distance to every goal in a set of candidate goals given a sequence of observations. Recently, \citeauthor{PereiraOrenMeneguzzi2017}~\shortcite{PereiraOrenMeneguzzi2017} develop landmark-based approaches for goal recognition, more specifically, they develop a two fast and accurate heuristics for goal recognition. 
Their first approach, called Goal Completion Heuristic, computes the ratio between the number of achieved landmarks and the total number of landmarks for a given candidate goal. 
The second approach, called Uniqueness Heuristic, uses the concept of landmark uniqueness value, representing the information value of the landmark for a particular candidate goal when compared to landmarks for all candidate goals. 
Thus, the heuristic estimative provided by this heuristic is the ratio between the sum of the uniqueness value of the achieved landmarks and the sum of the uniqueness value of all landmarks of a candidate goal. 

\section{Conclusions and Discussion}
\label{sec:conclusions}

We developed an approach for goal recognition in latent space using an LSTM network, obviating the need for human engineering to create a task for goal recognition. 
Other approaches require either human engineered domains~\cite{PereiraOrenMeneguzzi2017}, or an extensive amount of domain knowledge to build a PDDL domain~\cite{amado2018}.
Empirical evaluation on multiple datasets shows that while we can solve some problems with the same or higher accuracy than hand-coded problems, our LSTM approach does not easily generalize for goals outside the training dataset. 
Nevertheless, our approach provides a meaningful initial step towards goal recognition without human domain engineering and minimal amount of training data.
In summary the advantages of using our LSTM approach to recognize goals are: high accuracy and fast prediction time when dealing with known goals; no false positive predictions, given that it only predicts a single goal; no need of a PDDL domain, which requires extensive domain knowledge.
However, our approach has the following disadvantages: like most pure machine learning approaches, performance is tied to the robustness of the training dataset; requires training, which is unnecessary for classical goal recognition approaches; very limited generalizability with small datasets.


As future work, we aim to develop a dataset to test the ability of the different goal recognition approaches in latent space when dealing with noisy observations. 
In the LSTM case, a noisy observation would be a state in the encoded trace that is not relevant to achieving the desired goal (\idest, an unnecessary step performed by the agent being observed). 
Since \citeauthor{amado2018}~\shortcite{amado2018} compute every transition of the domain to generate a complete PDDL domain, we would like to investigate ways to use such information to improve the LSTM performance when dealing with goals that are not contained in the training set. Furthermore, we would like to study ways to improve generalization in our approach.
We envision using a percentage of the encoded transitions as a pre-training mechanism for the network, forcing the network to reconstruct many of the goals of the domain. 

To improve the quality of the datasets generated for the LSTM, we intend to use a top-$k$ planner~\cite{TOPK_katz_etal_icaps18}, which provides the ability to return $k$ distinct plans from a particular planning problem, being useful to generate datasets with both optimal and sub-optimal plans. This approach could improve the accuracy of the LSTM network when dealing with noisy and sub-optimal plans.


\bibliographystyle{named}
\bibliography{pal}

\end{document}